\documentclass[letterpaper]{article} 
\usepackage{aaai25}  
\usepackage{times}  
\usepackage{helvet}  
\usepackage{courier}  
\usepackage[hyphens]{url}  
\usepackage{subcaption}
\usepackage{float}
\usepackage{mathrsfs}
\usepackage{tabularx}
\usepackage{xcolor}

\usepackage{graphicx} 
\usepackage{natbib}  
\usepackage{caption} 
\usepackage{algorithm}
\usepackage{algorithmic}
\usepackage{newfloat}
\usepackage{listings}
\DeclareCaptionStyle{ruled}{labelfont=normalfont,labelsep=colon,strut=off} 
\floatstyle{ruled}
\newfloat{listing}{tb}{lst}{}
\floatname{listing}{Listing}
\title{Towards Physics-Guided Foundation Models\vspace{-1.25em}}
\author {
    Majid Farhadloo\textsuperscript{\rm 1}\equalcontrib,
    Arun Sharma\textsuperscript{\rm 1}\equalcontrib ,
    Mingzhou Yang\textsuperscript{\rm 1},
    Bharat Jayaprakash\textsuperscript{\rm 2},
    William Northrop\textsuperscript{\rm 2},
    Shashi Shekhar\textsuperscript{\rm 1}
}
\affiliations {
    \textsuperscript{\rm 1}Department of Computer Science, University of Minnesota, Twin Cities, USA\\
    \textsuperscript{\rm 2}Department of Mechanical Engineering, University of Minnesota, Twin Cities, USA\\

    \{farha043, sharm485, yang7492, jayap015, wnorthro, shekhar\}@umn.edu
}
\usepackage{bibentry}
\begin{document}
\vspace{-5em}
\maketitle
\vspace{-5em}
\begin{abstract}
Traditional foundation models are pre-trained on broad datasets to reduce the training resources (e.g., time, energy, labeled samples) needed for fine-tuning to a wide range of downstream tasks. However, traditional foundation models struggle with out-of-distribution prediction and can produce outputs that are unrealistic and physically infeasible.  We propose the notation of physics-guided foundation models (PGFM), that is, foundation models integrated with broad or general domain (e.g., scientific) physical knowledge applicable to a wide range of downstream tasks.
\end{abstract}
\vspace{-1.5em}
\section{Introduction}
Driven by the availability of large-scale datasets, advancements in computational power, and innovations in deep learning architectures, traditional foundation models (FMs) have significantly advanced the field of artificial intelligence \cite{chen2020simple, vaswani2017attention}. For a comprehensive survey on FMs and their history, interested readers may refer to \cite{zhou2024comprehensive}.



\textbf{Limitations of Foundation Models: }  Despite their versatility, purely data-driven FMs exhibit major limitations in scientific and engineering domains. One is their struggle with out-of-distribution situations, in which their benefits to downstream applications may be limited. This shortcoming is particularly problematic in specialized tasks that require nuanced understanding and adaptation to specific domain knowledge, such as climate science and healthcare. For instance, the Geospatial Foundation Model (Prithvi) \cite{b6}, trained exclusively on satellite imagery data from Landsat and Sentinel with multi-spectral bands, may fail to generalize to other data types, such as hyperspectral imagery or newer satellites with different spectral resolutions.  Second, purely data-driven foundation models (FMs) often violate fundamental physical principles such as energy conservation and motion dynamics. Without physics-based constraints, these models produce unrealistic and physically infeasible outputs. For example, models trained to estimate energy consumption and generate velocity profiles using transportation data such as onboard diagnostics and trajectory datasets may display rapid and unrealistic variations in velocity. Figure \ref{fig1}(a) shows velocity profiles without a jerk penalty that exhibit sharp peaks and drops, which reflect abrupt speed changes not typical in realistic driving scenarios. Figure \ref{fig1}(b) demonstrates that omitting a jerk penalty results in excessive jerk values far beyond passenger comfort thresholds \cite{de2023standards}.
Third, their black-box nature reduces transparency and makes predictions difficult to interpret \cite{b12}. These challenges highlight the need for Physics-Guided Foundation Models (PGFMs), which integrate scientific laws and physical constraints to enhance prediction reliability, robustness, and domain trust.


\textbf{Contributions.} Our primary contributions include the following. We formally define the concept of Physics-Guided Foundation Models (PGFM) and list a few methods to incorporate broad-domain physical knowledge. Additionally, we provide concrete examples illustrating the limitations of existing FMs and the necessity of PGFM.

\vspace{-1em}
\section{Vision}\label{sec2}
A physics-guided foundation model is a foundational model incorporating broad domain knowledge and a general understanding of the domain’s fundamental concepts and principles. For example, popular knowledge graphs and taxonomies \cite{ji2021survey} may strengthen foundation models for a broad range of downstream applications. Also, conservation laws (e.g., mass, energy) apply to a broad range of downstream physical science tasks. Concepts of speed, acceleration, and laws of motion are applicable to a broad range of tasks related to moving objects. Maxwell's laws are also widely applicable to tasks related to electromagnetism. A generalizable model of biogeochemistry may be used as a PGFM with extensive data training.

\begin{figure}[h]
\centering
\begin{subfigure}[b]{0.55\linewidth} 
    \centering
    \includegraphics[width=\linewidth]{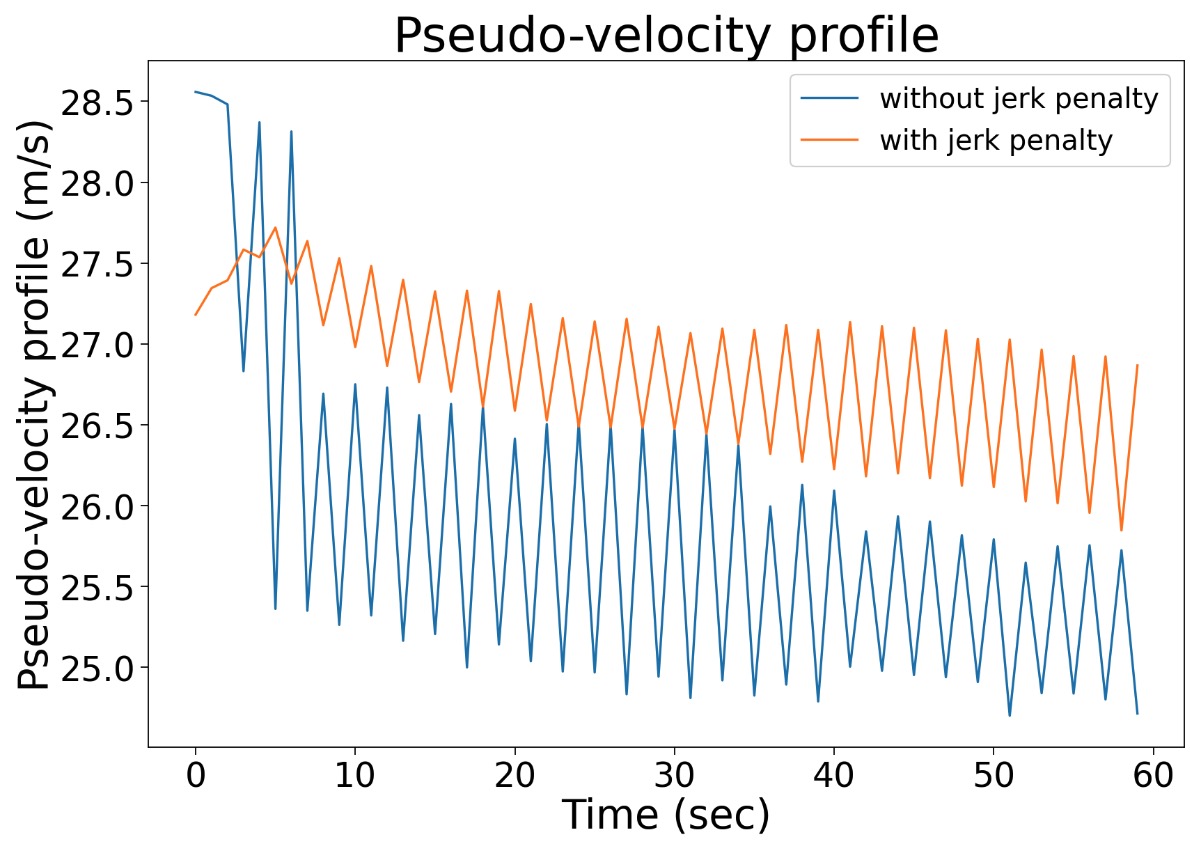}
    \vspace{-1.5em}
    \caption{Velocity profiles}
    \label{fig:subfig1}
\end{subfigure}
\hfill 
\begin{subfigure}[b]{0.55\linewidth} 
    \centering
    \includegraphics[width=\linewidth]{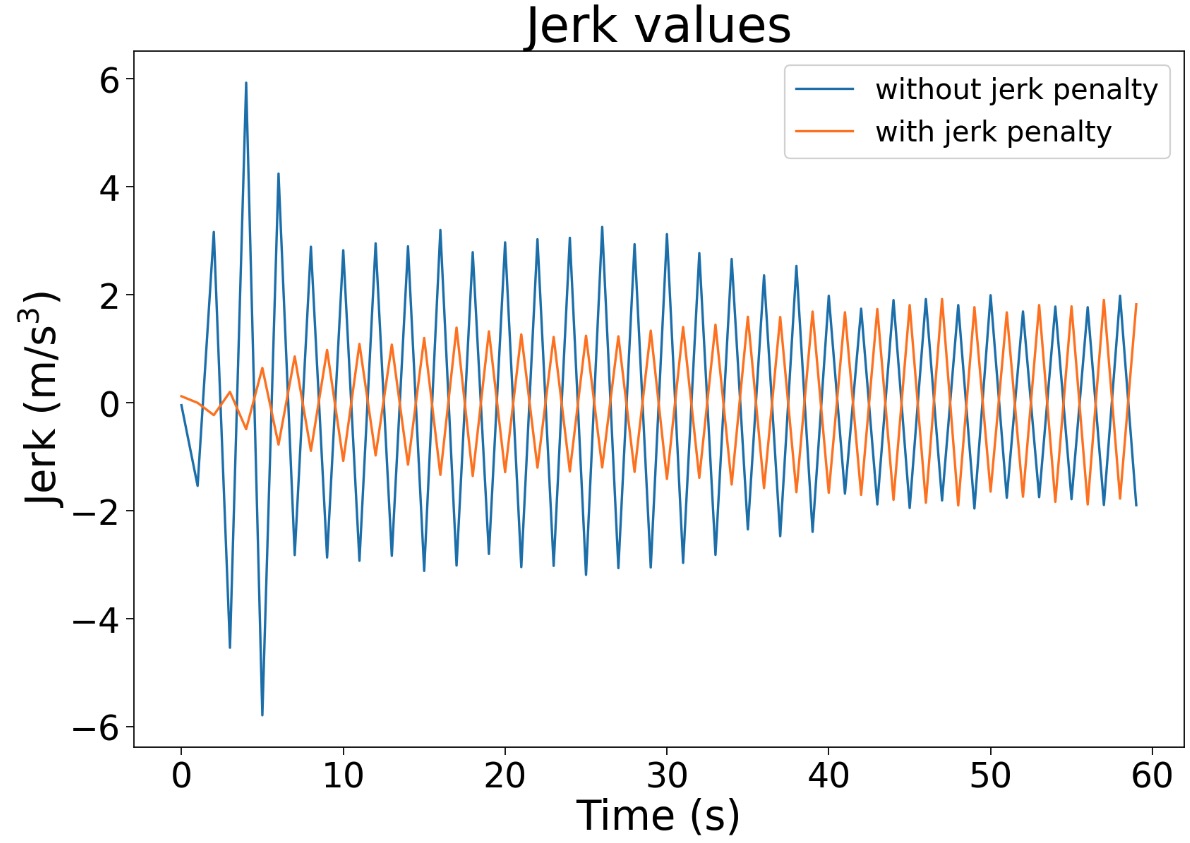}
    \vspace{-1.5em}
    \caption{Jerk values}
    \label{fig:subfig2}
\end{subfigure}
\vspace{-1em}
\caption{Neural network-generated velocity profiles and corresponding jerk values.}
\label{fig1}
\vspace{-2em} 
\end{figure}

A variety of approaches can be employed to embed physical knowledge into foundation models. \textit{Physics-constrained learning} enforces real-world constraints by introducing domain rules into the learning process through the loss function or regularization that safeguards model predictions to ensure physical feasibility (e.g., \cite{b13}). Similarly, \textit{architecture-level integration} (e.g., \cite{b13, jia2019physics}) incorporates domain principles directly into model structures, aligning network design with physics-based rules and patterns. In recent work \cite{b13}, we demonstrated that incorporating a physical constraint, such as a jerk penalty, in velocity profiling reduced the sharp fluctuations common in unconstrained models, as shown in Figure \ref{fig1}. This adjustment produced smoother outputs that better reflect realistic physical behaviors and practical driving dynamics.

The effectiveness of these approaches often depends on the type of physical knowledge embedded into the model. Wide physics knowledge refers to universal principles, such as conservation of energy or laws of motion, that apply across a broad range of scenarios. These principles are foundational and provide generalizable representations during pretraining. In contrast, narrow physics knowledge consists of task-specific principles tailored to specialized applications, such as aerodynamic drag models for vehicle energy consumption or battery discharge dynamics. For instance, Peukert’s law \cite{doerffel2006critical}, which describes the nonlinear discharge of a battery under varying current loads, is highly specific to certain powertrain types and conditions. Instead of embedding this directly in pretraining, broader energy dynamics, such as the relationship between energy use and current, can be incorporated to develop transferable representations. Narrow principles like Peukert’s law then become relevant during fine-tuning for downstream tasks, such as estimating the remaining driving range of an electric vehicle under specific operational constraints.

This distinction between wide and narrow physics knowledge is also evident in the datasets used to train Physics-Guided Foundation Models (PGFMs). Pretraining typically requires large-scale, diverse datasets that reflect broad physical principles. Examples include traffic flow datasets from highway cameras, vehicle trajectory datasets such as NGSIM \cite{coifman2017critical}, and energy datasets like FASTSim simulations \cite{brooker2015fastsim}, which are calibrated to US Environmental Protection Agency standards. These datasets enable the model to generalize fundamental relationships, such as energy conservation and speed-density dynamics. For downstream tasks, more specialized datasets tailored to specific applications are necessary. For example, aerodynamic drag models may be refined using vehicle-specific wind tunnel data, while battery discharge predictions may be fine-tuned with high-resolution battery management system data that capture real-world Peukert’s law behavior.

The distinction between wide and narrow knowledge aligns closely with the broader differences between PGFMs and other machine learning paradigms, as summarized in Table \ref{tab:comparison}. Task-specific models, trained on narrow datasets, are limited in their ability to generalize across tasks and rarely integrate scientific or domain knowledge. Foundation models (FMs), while trained on broad datasets, often remain purely data-driven and lack integration of physical principles, which constrains their applicability to physics-informed domains. Physics-Guided Task-Specific (PGTS) models focus on integrating deep but narrowly defined domain knowledge into the learning process \cite{karpatne2017theory, daw2022physics}, producing highly specialized models that excel in specific tasks. PGFMs combine the strengths of these approaches by integrating broad training data with wide physics knowledge, enabling them to generalize across tasks, while leveraging narrow domain principles during fine-tuning to support task-specific requirements.
\vspace{-.8em}

\begin{table}[h]
\setlength{\abovecaptionskip}{2pt} 
\setlength{\belowcaptionskip}{2pt} 
\setlength{\intextsep}{2pt} 
\renewcommand{\arraystretch}{0.9} 
\captionsetup{skip=2pt} 
\caption{Comparison of PGFMs with Related Concepts}
\centering
\scriptsize
\begin{tabular}{|p{0.22\columnwidth}|p{0.15\columnwidth}|p{0.21\columnwidth}|p{0.22\columnwidth}|}
\hline
\textbf{Technique} & \textbf{Training Data} & \textbf{Scientific Knowledge} & \textbf{Downstream Tasks Range}  \\
\hline
Task-specific Model & -- & -- & Narrow \\
\hline
FMs & Broad & --  & Wide\\
\hline
PGTS & -- & Narrow & Narrow \\
\hline
PGFMs & Broad & Wide & Wide \\
\hline
\end{tabular}
\label{tab:comparison}
\vspace{-1em}
\end{table}

\vspace{-1em}
\section{Conclusion and Future Work}\label{sec5}
PGFM enhances foundation models by systematically integrating domain-specific physcial knowledge, improving performance, robustness, and interpretability for diverse applications. Future research offers promising advancements in AI across fields like healthcare \cite{farhadloo2024spatial}, geospatial analysis, and engineering. We plan to develop and evaluate a PGFM model, focusing on domain knowledge integration using metrics like sample complexity. 

Advancing PGFM in future work will involve developing hybrid frameworks that combine the adaptability of retrieval-augmented generation \cite{lewis2020retrieval} with real-time updates to optimize domain adaptation and reduce training resource demands. We will also explore PGFMs in the context of recent multi-modal foundation models \cite{ravirathinam2024towards}, focusing on integrating physical knowledge \cite{sharma2024physics, sharma2022analyzing, sharma2022towards} from diverse input sources and its impact on training procedures and downstream applications. We will also explore additional criteria for comparing PGFMs with other machine learning models, including training methods and challenges such as model complexity, computational requirements, location dependency \cite{yang2025climate, ghosh2024towardssig, ghosh2024towardsarxiv, ghosh2024reducing, ghosh2024towardscosit, ghosh2023reducing, ghosh2022towards}, and calibration needs \cite{farhadloo2024towards, farhadloo2024spatial}. In the long term, we aim to address challenges such as explainability \cite{arrieta2020explainable, farhadloo2022samcnet}, robustness, bias, data quality, and adaptability. 

\clearpage
\section*{Acknowledgments}{This material is based on work supported by the USDA under Grant No. 2023-67021-39829, the National Science Foundation under Grant No. 1901099, the USDOE Office of Energy Efficiency and Renewable Energy under FOA No. DE-FOA0002044, and USDA under Grant No. 2021-51181-35861. We also thank Kim Kofolt and the Spatial Computing Research Group for their valuable comments and contributions.}
\bibliography{aaai25}

\end{document}